\documentclass{INTERSPEECH2023}

\interspeechcameraready

\usepackage{tabularx}
\usepackage{multirow}

\newcommand{\mytable}{
	\centering
	\renewcommand{\arraystretch}{1.2}
}
\newcommand{\captionsep}{\vspace*{-5pt}}
\newcolumntype{L}{>{\raggedright\arraybackslash}X}

\title{Towards hate speech detection in low-resource languages:\\ Comparing ASR to acoustic word embeddings on Wolof and Swahili}

\name{Christiaan Jacobs$^1$, Nathana\"el Carraz Rakotonirina$^{2,5}$, Everlyn Asiko Chimoto$^{3,4}$, \\ Bruce A. Bassett$^{3,4,5}$, Herman Kamper$^1$}

\address{ \normalsize
  $^1$Department of Electrical \& Electronic Engineering, Stellenbosch University, South Africa \\ 
  $^2$Universitat Pompeu Fabra, Spain \quad 
  $^3$African Institute for Mathematical Sciences, Cape Town, South Africa \\
  $^4$Department of Mathematics \& Applied Mathematics, University of Cape Town, South Africa \\
  \quad 
  $^5$VoxCroft Analytics, Cape Town, South Africa (http://www.voxcroft.com)}

\usepackage{cite}
\usepackage{microtype}

\usepackage{xcolor}
\definecolor{mycolor}{HTML}{FF6600}

\begin{document}

\maketitle

\begin{abstract}
We consider hate speech detection through keyword spotting on radio broadcasts. One approach is to build an automatic speech recognition (ASR) system for the target low-resource language. We compare this to using acoustic word embedding (AWE) models that map speech segments to a space where matching words have similar vectors. We specifically use a multilingual AWE model trained on labelled data from well-resourced languages to spot keywords in data in the unseen target language. In contrast to ASR, the AWE approach only requires a few keyword exemplars. In controlled experiments on Wolof and Swahili where training and test data are from the same domain, an ASR model trained on just five minutes of data outperforms the AWE approach. But in an in-the-wild test on Swahili radio broadcasts with actual hate speech keywords, the AWE model (using one minute of template data) is 
more robust, giving similar performance to an ASR system trained on 30 hours of labelled data.
\end{abstract}

\section{Introduction}
\label{sec:intro}

Hate speech is a pervasive problem in many parts of the world.
In some {developing} countries, radio broadcasts are the primary medium for communicating news and information to the public~\cite{chalk_hate_1999}.
They also serve as a platform for public discourse, enabling freedom of expression~\cite{somerville_violence_2011}.
Monitoring hate speech on these radio broadcasts in low-resource languages is extremely challenging due to factors such as background noise, speaker variability, and the presence of accents or dialects {in the audio}. 

Hate speech detection can be {supported by using} a keyword spotting (KWS) system {to search through an audio corpus for} a predetermined set of keywords {indicative} of hate speech.
The conventional {KWS} approach involves transcribing the audio corpus with an automatic speech recognition (ASR) {system} and then {searching for keywords in the output.}
However, for low-resource languages, the absence of large quantities of transcribed speech can make it difficult to train high-quality ASR models.
Several studies have looked into this problem{, proposing methods to train models on limited amounts of labelled data}~\cite{chen_small-footprint_2014, audhkhasi_end--end_2017, menon_fast_2018, mekonnen_end--end_2022}.

{When we want to rapidly develop and deploy a KWS system for hate speech detection in a new setting, it might be better to opt for an ASR-free approach.}
The most popular {methodology} extends query-by-example (QbE), which uses a {\textit{spoken}} instance of a keyword to search through an audio corpus.
{To use QbE for KWS, we need a small number of spoken templates to serve as queries for the keywords of interest; this is the approach followed in~\cite{menon_fast_2018, van_der_westhuizen_feature_2022}.}
Dynamic time warping is typically used to match {the} speech features of a spoken query to a search utterance~\cite{park_unsupervised_2008, hazen_query-by-example_2009, zhang_unsupervised_2009, jansen_indexing_2012}.

Recent alternative approaches jointly map search segments and query segments to a fixed vector space{, allowing for faster comparisons}.
Various neural networks have been considered to obtain these {\textit{acoustic word embeddings} (AWEs)}~\cite{chung_unsupervised_2016, kamper_deep_2016, settle_discriminative_2016, settle_query-by-example_2017, kamper_truly_2019, abdullah_acoustic_2021}.
{One approach to {quickly} obtain {robust} embeddings for a new language is to train a multilingual AWE model} on labelled data from multiple well-resourced languages, before applying it to {the} target low-resource language~\cite{ram_neural_2019, hu_multilingual_2020, kamper_improved_2021, jacobs_acoustic_2021, jacobs_multilingual_2021}.
{Although these multilingual AWE models have proven successful in controlled experiments,} there has been limited work investigating the effectiveness of these systems beyond the experimental environment---{and} even less {so} for hate speech detection.

In this paper, we investigate hate speech detection {in low-resource languages} through KWS.
{Our main goal is to compare ASR to a multilingual AWE-based approach on real data.}
Specifically, we {consider} the performance of {different ASR and AWE-based} systems on {real} Swahili radio broadcast data scraped from stations in Kenya, a country in sub-Saharan Africa.
{For our ASR systems, we build on recent} advancements in multilingual speech recognition models, {enabling a} pre-trained ASR model {to be} fine-tuned using much less training data {than that required for training an ASR model from scratch}~\cite{baevski_wav2vec_2020, zhao_improving_2022, nguyen_improving_2023}.
{We} investigate the trade-off between the amount of  {ASR} training data and KWS performance.
We compare the fine-tuned ASR models {to} a KWS system using QbE with multilingual AWEs~\cite{jacobs_multilingual_2021, hu_acoustic_2021}.

{Building up to our final experiments on real radio broadcasts, we first perform experiments} in a controlled environment {where training and test data come from the same domain.}
Given that the pre-trained ASR model used in our experiments includes Swahili {as a pre-training language}, we {also consider performance on} another low-resource language, Wolof, in our {controlled} 
experiments.
Then, we put the ASR and AWE KWS systems to a real-life test by evaluating performance on Swahili radio broadcasts.
{In the controlled experiments we find that} fine-tuning an ASR model using as little as five minutes of labelled data outperforms the AWE{-based} KWS system which relies on roughly one minute of template audio.
{However, in the out-of-domain experiment on real scraped Swahili} radio broadcasts, the AWE system proves to be more robust by almost reaching the performance of an ASR model fine-tuned on 30 hours of data. 
Our findings suggest that KWS using multilingual AWEs is a promising approach for quickly implementing hate speech detection in a new unseen language {if resources are severely limited}.
\section{Towards hate speech detection through low-resource keyword spotting}
\label{sec:kws}

In this section, we briefly give the overall methodology that we follow towards our end goal of hate speech detection on Swahili radio broadcasts.
Exact details are then given in what follows.

We start by developing Wolof and Swahili KWS systems in a controlled environment where we use parallel audio-text data from experimental datasets to train and test our models.
These datasets~\cite{gauthier_collecting_2016, ardila_common_2020} contain high-quality, noise-free read speech.
We then test our models on data from these corpora.
We refer to these experiments as \textit{in-domain} because the test data comes from the same domain as the training data.
In this controlled setup, we perform KWS using a set of keywords drawn from development data.
These are not necessarily hate speech keywords.
The benefit of this controlled setting is that we have transcriptions for the test data, and can therefore quantify performance exactly.

We then apply the models to the \textit{out-of-domain} Swahili radio broadcasts.
These recordings were obtained from three radio stations in Kenya.
They consist of a mix of read and spontaneous speech, and also include non-speech audio like music and advertisements.
Additionally, there may be instances of different accents and languages present within the recordings.
This is the type of real-world audio that we would like to monitor for hate speech using KWS.
As our keywords in this context, we utilise actual hate speech keywords.
These were labelled as inflammatory words by native Swahili experts familiar with the media environment.
We should emphasise that, when we detect occurrences of these words, they might not necessarily respond to hate speech (e.g.\ the English sentence ``do not kill your neighbour'' is not hate speech despite containing the word ``kill'').
In a deployed system the utterances flagged by our KWS approach would be passed on to human moderators for further review and hate speech assessment.
In this work we also use a human moderator, but only to mark whether a detected keyword actually occurred in an utterance. We do this for our in-the-wild test since we do not have transcriptions for the radio broadcasts.
\section{Keyword spotting with ASR}
\label{sec:asr}

ASR is one of the most straightforward approaches to hate speech detection via {KWS} in low-resource languages. Given a set of keywords, the ASR model first transcribes the audio into text, then the keywords are searched for in the resulting text~\cite{saeb+etal_interspeech2017}.
One property of this approach is that the downstream KWS performance depends on the ASR model's performance.

We use an ASR model from the Wav2vec 2.0 XLS-R family~\cite{baevski_wav2vec_2020, babu_xls-r_2022}.
XLS-R is a large-scale cross-lingual speech model pre-trained on half a million hours of unlabelled speech data in 128 languages (including Swahili but not Wolof).
The speech representation learned during cross-lingual pre-training improves performance for low-resource languages by leveraging data from the high-resource languages on which it is pre-trained.
Using the pre-trained model as a starting point, the model is fine-tuned on labelled data from the language of interest.

Previous work has shown that this fine-tuning approach leads to competitive ASR performance even with limited amounts of labelled data~\cite{babu_xls-r_2022}.
In our experiments (Table \ref{tbl:awe} specifically) we show that the sample efficiency of the XLS-R model also extends to KWS when training on just five minutes of labelled data.
\section{Keyword spotting with multilingual acoustic word embeddings}
\label{sec:awe}

For a severely under-resourced language, there might not even be limited amounts of labelled data available to fine-tune a pre-trained ASR model.
In this scenario, collecting only a few spoken instances of a keyword of interest would allow for KWS through query-by-example (QbE) search~{\cite{menon_fast_2018, van_der_westhuizen_feature_2022}}.
QbE is the task of retrieving utterances from a search corpus using a spoken instance of a keyword, instead of a written keyword.

\begin{figure}[!t]
    \centering
    \includegraphics[width=0.95\linewidth]{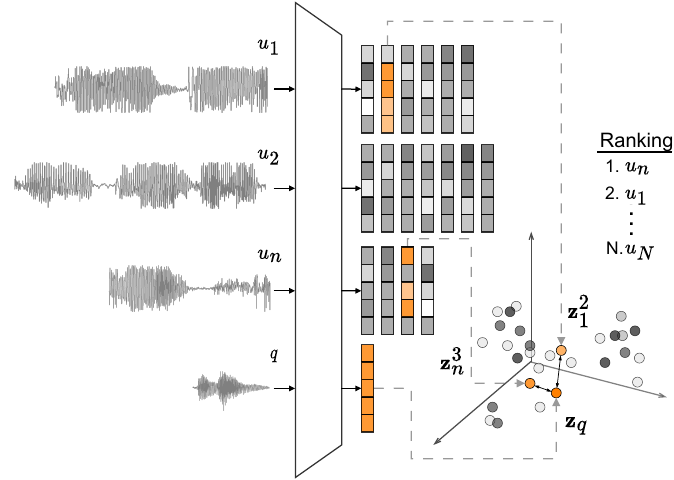}
    \vspace{-2mm}
    \caption{{Search utterances} ($u_1, u_2, \ldots, u_N$) are segmented and mapped to a fixed-dimensional space, along with a query segment $q$, to produce embeddings $\mathbf{z}_n^i$ and $\mathbf{z}_q$, respectively.
    {Subsequently,} search utterances are ranked based on the shortest distance between a search segment $\mathbf{z}_n^i $ and the query vector $\mathbf{z}_q$.}
    \label{fig:qbe_awe}
    \vspace{-3mm}
\end{figure}

The workings of an AWE-based QbE system are displayed in Figure~\ref{fig:qbe_awe}.
First, the search corpus is segmented into word-like segments.
Ideally, we would like to use true word boundaries to segment each utterance in the search collection.
However, in an under-resourced setting, word boundaries are not available.
Therefore, we follow an exhaustive segmentation technique where utterances are split into overlapping segments of some minimum to some maximum duration~\cite{kamper_semantic_2019}.
Concretely, utterances from the search corpus ($u_1, u_2, \ldots, u_N$) are segmented to produce a set of variable-length speech segments $S = \{s_n^i\}$, where $i$ is the $i^\text{th}$ search segment extracted from utterance $n$.

A multilingual AWE model is then applied to the segmented search corpus and query segment, mapping them to the same fixed-dimensional vector space.
We specifically use the CAE-RNN architecture~\cite{kamper_improved_2021, jacobs_acoustic_2021} to train a multilingual AWE model.
This model is trained using an encoder-decoder {recurrent neural network (RNN)} structure, where the encoder takes a variable-length word segment as input to produce a fixed-dimensional output. 
The output vector of the encoder is then fed as input to a decoder RNN, which reconstructs a target that is not identical to the input but rather an instance of the same word type.
This model is trained on word pairs using labelled data from multiple languages (but not the target language).
At inference time, we take the projection of the final encoder RNN hidden state as the AWE of an input segment.

Applying the AWE model to search segments $S$ and query segment $q$ produces search embeddings $\mathbf{z}_n^i$ and query embedding $\mathbf{z}_q$, respectively.
Utterances in the search corpus can then be ranked based on the segment-query distances $d(\mathbf{z}_n^i, \mathbf{z}_q)$.
We can then predict that all query-search pairs with a distance below some threshold are positive matches (this requires a threshold to be tuned on development data).
With these predictions we can calculate common KWS metrics and compare them to those achieved by the ASR KWS systems.
\section{Experimental setup}
\label{sec:setup}

\textbf{Data.} We train and evaluate our models on Swahili and Wolof data, collected from the Common Voice~\cite{ardila_common_2020} and ALFFA~\cite{gauthier_collecting_2016} datasets, respectively.
For both datasets, we use the default train, development, and test splits.
In our controlled experiments, we evaluate in-domain KWS performance on the Swahili and Wolof test splits. 
The test split of both languages serves as the search collection and is not used during model development.
The Swahili search collection contains 8~941 utterances and has a duration of approximately 14 hours, while the Wolof set contains 2~000 utterances and has a duration of approximately 2.5 hours. 

We perform our in-the-wild test on recorded radio broadcasts from three different radio stations in Kenya.
We do not have any transcriptions, speaker information or content information on these audio clips.
We segment this data using a diarisation system that also attempts to remove music.
Apart from this preprocessing step, we simply apply our systems to this out-of-domain data without further changes or calibration.
After this preprocessing step, 19\,716 utterances are obtained, which is our out-of-domain search collection.
The duration of all utterances is between three and 30 seconds.
As explained below, a human expert is used to evaluate the system's performance on this set.

\textbf{ASR model.}
For the ASR KWS systems (Section~\ref{sec:asr}), we control the amount of training data and train three ASR models for Wolof and Swahili.
Specifically, we fine-tune the XLS-R model~\cite{babu_xls-r_2022} for each language on 30 hours, one hour, and five minutes of labelled training data.

\begin{table}[!b]	
	\mytable
        \vspace{-2mm}
	\caption{Swahili hate speech keywords used for in-the-wild KWS on Swahili radio broadcasts, with {their} English translations.}
	\addtolength{\tabcolsep}{10pt}
	\captionsep
	\eightpt
	\begin{tabularx}{1\linewidth}{@{}ll|ll@{}}
		\toprule				
		Swahili & English & Swahili & English \\ [0.5mm]
		\midrule
		vita & war & wezi & thieves \\
		damu & blood & majimbo & states \\
		hama  & move & wakora & conmen  \\
		kabila & tribe & panga & machete \\
		utapeli & fraud & takataka & garbage \\
		mende & cockroaches &  mjinga & stupid \\
		kitendawili & riddle & fala & stupid\\
		\bottomrule		
	\end{tabularx}
	\label{tbl:keywords}
\end{table}

\textbf{AWE model.} 
For the AWE-based KWS system (Section~\ref{sec:awe}), we train a multilingual AWE model on five different Common Voice languages: Abkhazian (ab), Czech (cs), Basque (eu), Swedish (sv), Tamil (ta).
Word boundaries are obtained using the Montreal forced-aligner~\cite{mcauliffe_montreal_2017}.
We pool the data from all languages and extract 300k positive word pairs which we use to train a multilingual CAE-RNN AWE model~\cite{kamper_improved_2021, jacobs_acoustic_2021}.
The encoder and decoder each consist of three unidirectional RNNs with 400-dimensional hidden vectors and an embedding size of 100 dimensions.  
{As inputs to the encoder, we} use XLS-R to extract speech features; we use the output of the $12^\text{th}$ transformer layer, producing input features with $1024$ dimensions.
Preliminary experiments showed using these features outperforms mel-frequency cepstral coefficients in AWE modelling.
We employ Adam optimisation~\cite{kingma_adam_2017} with a learning rate of $0.001$.

\textbf{Controlled evaluation.} To measure in-domain KWS performance, we apply the ASR and AWE-based KWS systems to the same search corpora.
Each utterance is labelled as either 1 (if the keyword is present) or 0 (if the keyword is not present).
We report precision, recall and $F_1$-score.
For the ASR KWS systems, we simply {check whether a keyword occurs in the predicted transcript.}
For the AWE-based KWS, sub-segments for utterances in the search collection are obtained by extracting windows ranging from 20 to 35 frames with a 5-frame overlap and then applying an AWE model to each segment.
We use a set of ten templates of 36 unique keywords with a minimum character length of five for Swahili, and 15 unique keywords with a minimum character length of four, for Wolof.
These keywords were randomly sampled from a larger set of words that appear at least ten times in both the development and test sets.
Query templates are drawn from development data, embedded, and averaged to obtain a single AWE embedding representing a keyword.
The similarity between a query and search segment is calculated using cosine distance.
The threshold value in the AWE-based KWS is tuned for the highest $F_1$-score across all keywords on the controlled test data.

\textbf{In-the-wild evaluation.} For the in-the-wild KWS, we apply the systems directly to the out-of-domain Swahili radio broadcasts.
We use a set of keywords labelled as inflammatory by expert analysts familiar with the media environment for the purpose of hate speech detection.
These keywords with their English translations are given in Table~\ref{tbl:keywords}.
For the AWE-based KWS system, ten query templates per keyword are extracted from the in-domain Swahili data.
We evaluate KWS performance by asking each approach to give 100 utterances out of the search collection that are most likely to contain any of the hate speech keywords.
In the absence of per-word confidence scores for the ASR models, here we simply take 100 random utterances that were predicted to contain a keyword (for most models, the total number of utterances was around 150).

For the AWE-based KWS, we use the 100 highest-ranked utterances. Because we do not have transcriptions for this data, we provide a native Swahili speaker with untranscribed recordings of the utterances that were predicted to contain a keyword. They then mark whether the keyword was indeed present.
\section{Results}
\label{sec:results}

\begin{table}[!b]	
	\mytable
         \vspace{-2mm}    
	\caption{
		ASR and AWE KWS results (\%) on in-domain test data. 
        For the ASR systems, the XLS-R model is fine-tuned on each target language, controlling the amount of training data.
        For the AWE KWS system, a supervised multilingual AWE is trained on multiple well-resourced languages and applied to the two low-resource target languages.}
	\captionsep
	\eightpt
	\begin{tabularx}{\linewidth}{@{\extracolsep{-6pt}}L*{6}{m{6.5mm}}}
		\toprule				
		\multirow{4}{1cm}{Model} & \multicolumn{3}{c}{Swahili} & \multicolumn{3}{c}{Wolof} \\
		\cmidrule(r){2-4} 
		\cmidrule(l){5-7}
		
		& Prec. &  Rec. & $F_1$ & Prec. &  Rec. & $F_1$  \\ 
		\midrule
				
		\underline{\textit{ASR}:} & & &  & & \\ [1mm]
  		XLS-R (30-h) & 97.6 & 98.6 &  {98.0} &  93.1 & 87.1 & 90.0  \\
		XLS-R (1-h) &96.4 & 76.1 & {85.1}  & 93.1 & 74.3 & 82.7  \\
		XLS-R (5-min) 
        & 95.4 & 42.6 &  58.9  & 88.0 & 45.6 & 60.0  \\

        \addlinespace

        \underline{\textit{Multilingual AWE}:} & & & & & & \\[1mm]
            CAE-RNN (ab+cs+eu+sv+ta)  & 57.1 &  56.1 & 56.6 & 44.0 & 58.4 & 50.2 	\\
            
            \bottomrule		
	\end{tabularx}
	\label{tbl:asr}
\end{table}

\subsection{KWS results in a controlled environment}
\label{ssec:results_in}
\vspace{-1mm}

We first look at results in a controlled {test} on in-domain Swahili and Wolof data.
The first three lines in Table~\ref{tbl:asr} report the ASR KWS results.
The results show that fine-tuning XLS-R on only five minutes of labelled data achieves decent KWS performance, with $F_1$-scores of {59\%} and 60\% for Swahili and Wolof, respectively.
{The multilingual AWE system using roughly one minute of template data, performs worse on most metrics.
The only metric on which the AWE system is better is recall, where it achieves higher scores than the 5-minute ASR system on both languages.}
{In practice, the AWE system could therefore be a better option if recall is important.}

{We briefly consider how ASR performance is affected by the amount of training data.}
{For reference, the word error rates (WERs) for the} 
30-hour, 1-hour, and 5-minute models {are respectively} 9\%, 36\%, and 62\% {on Swahili.}
For Wolof, the WERs are 27\%, 44\%, and 68\%.
{As expected, more training data gives better ASR and KWS scores. But it is noteworthy that with just five minutes of training data, we can already spot keywords with high precision (95\% and 88\% for Swahili and Wolof, respectively). This is especially useful for our use case, where we want to rapidly develop KWS applications in severely low-resourced settings.}
{The other noteworthy finding from the ASR results is that scores} for Swahili are not notability higher than Wolof, although {the former is one of XLS-R's pretraining languages.}

\begin{table}[t]	
	\mytable
	\caption{
			Supervised AWE KWS results (\%) on in-domain Swahili test data.
            The supervised monolingual CAE-RNN AWE model is trained using labelled Swahili data.
            It is applied with two segmentation configurations: using ground-truth word boundaries (true segm.), and using a variable-length window which is swept across the search collection (random segm.). 
            Here we also report the standard QbE metrics P@10 and P@N. 
			}	
	\captionsep
	\eightpt
	\begin{tabularx}{1\linewidth}{@{\extracolsep{-4pt}}L*{5}{m{6mm}<{\centering}}}
		\toprule				
		Model & Prec. & Rec. & $F_1$ & P@10 & P@N  \\ 
		\midrule
		\underline{\textit{Supervised monolingual}:} & & & & &   \\ [1mm]
		CAE-RNN (true segm.) & 95.5 & 90.8 &{93.1} & 98.6 & 94.3  \\
		CAE-RNN (random segm.)  & 79.2 &  76.1 & {77.6} & 92.2 & 90.9 \\
  		\addlinespace
		\underline{\textit{Supervised multilingual}:} & & & & & \\ [1mm]
		CAE-RNN ab+cs+eu+sv+ta (random segm.)  & 57.1 &  56.1 & {56.6} & 87.8 & 64.7 \\
		\bottomrule		
	\end{tabularx}
	\label{tbl:awe}
 \vspace{-5mm}
\end{table}

{We now turn to the AWE-based approach; specifically, we ask what the upper bound on performance would be if we had more training data or a more idealised search setting.}
Therefore, for a moment, we assume we have labelled data available and perform top-line experiments on the in-domain Swahili data, shown in Table~\ref{tbl:awe}.   
A supervised Swahili AWE model is trained {on 30 hours of labelled data} and applied to the search collection that is segmented using true word boundaries.
{Results are shown in the first row of Table~\ref{tbl:awe}, serving} as the top-line performance for AWE KWS, with a high $F_1$-score of 93\%.
{Compared to the 30-hour ASR system (Table~\ref{tbl:asr}), the idealised AWE system comes closer to the $F_1$ of 98\%.}
The second row in Table~\ref{tbl:awe} shows KWS performance using a supervised Swahili AWE model but applied to search segments extracted by sliding a variable-length window across the search collection{---the way we apply the AWE approach in practice and in Table~\ref{tbl:asr}}.
Segmentation without {true} word boundaries incurs a significant penalty, resulting in an $F_1$-score drop to 77\%.
{It is clear that the sliding window approach has a large effect on downstream performance, so future work should consider more sophisticated unsupervised word segmentation techniques.}

\vspace{-1mm}
\subsection{KWS results in the wild}
\label{ssec:results_out}
\vspace{-1mm}
{We now turn to our main research question: comparing} ASR to multilingual AWE-based KWS on real-life, out-of-domain audio in a low-resource setting. 
As mentioned in Section~\ref{sec:setup}, systems {are}  applied to out-of-domain Swahili {radio} broadcasts, after which, for each system, the top 100 utterances predicted to contain a hate speech keyword (Table~\ref{tbl:keywords}) {are} manually reviewed.
The results are {given} in Table~\ref{tbl:wild}.

{The table reports precision: the proportion of retrieved top-100 utterances that correctly contain a hate speech keyword.}
Surprisingly, we see that the multilingual AWE KWS system {achieves} a precision of 45\%---better than the 5-minute and even the 1-hour ASR system {in this in-the-wild test}.
This is in contrast to the in-domain KWS results, where the 1-hour ASR model outperformed the AWE KWS system {(Table~\ref{tbl:asr})}.

{Further investigation is required to understand exactly why the relative performance of the ASR and AWE systems are affected differently when applied to out-of-domain data.
However, it is worth noting that several studies have shown that ASR system performance can drop dramatically when it is applied to data outside of its training domain~\cite{seltzer_investigation_2013, likhomanenko_rethinking_2021, hsu_robust_2021}.
One example where this can be seen in this case is how many of the ASR retrievals contain music (which we asked the human annotator to mark).}

The in-the-wild search collection has been diarised automatically, which included a step to remove music segments. But this preprocessing step is not perfect: the search collection still ends up with some music (which neither the ASR nor AWE systems have seen in training).
Table~\ref{tbl:wild} shows that, out of the top-100 utterances for the 30-hour ASR system, 30\% contained music. This decreases for the 1-hour (29\%) and 5-minute (17\%) ASR systems. But the AWE KWS system only retrieves two utterances containing music.
And of these two, one utterance actually did contain a hate speech keyword in the music lyrics (there are also examples of such correct matches in the ASR music retrievals). Nevertheless, this shows that the AWE approach seem to be more robust to domain mismatch compared to training an ASR system on one hour or five minutes of labelled~data.

\begin{table}[t]	
	\mytable
	\caption{
		In-the-wild KWS results (\%) on Swahili radio broadcasts.
		For each system, we report the percentage of utterances retrieved containing a keyword (precision) and the percentage of utterances retrieved containing music.
	}
	
	\captionsep
	\centering
	\eightpt
	\begin{tabularx}{\columnwidth}{@{\extracolsep{0pt}}L m{11mm}<{\centering}m{11mm}<{\centering}m{12mm}<{\centering}}
		\toprule

		Model & {Precision} & Music \\
		\midrule
		\underline{\textit{ASR}:} & &   \\ [1mm]
		XLS-R (30-h) & 52 & 30  \\
		XLS-R (1-h) & 42 &  29  \\
		XLS-R (5-min) & 36 & 17   \\
        \addlinespace
		\underline{\textit{Multilingual AWE}:} & &   \\ [1mm]
		CAE-RNN ab+cs+eu+sv+ta & 45 & 2  \\		  
		\bottomrule		
	\end{tabularx}
	\label{tbl:wild}
		\vspace{-4mm}
\end{table}

\section{Conclusion}
\label{sec:conclusion}

{This paper considered the problem of} keyword spotting (KWS) for the purpose of hate speech detection in low-resource languages.
We compare two KWS systems for {data from} Swahili and Wolof: a fine-tuned automatic speech recognition (ASR) model using different amounts of training data, and an ASR-free KWS system that utilises {multilingual} acoustic word embeddings (AWEs). 
The results show that fine-tuning a pre-trained multilingual ASR model using even a small amount of labelled data can outperform an AWE-based KWS system in controlled environments {where training and test data come from the same domain}.  
However, the AWE-based KWS system is more robust on out-of-domain radio broadcast data and achieves comparable results to an ASR model fine-tuned on 30 hours of labelled data. 
{In the end, the merits of ASR vs AWE KWS will come down to the practical setting: it will depend on whether labelled training data can be collected from the target domain or not, and whether precision or recall is more important.}

\vspace{-1mm}
\section{Acknowledgements}
UPF has received funding from the European Research Council (ERC) under the European Union's Horizon 2020 research and innovation programme (grant no.\ 101019291). We thank the VoxCroft Analytics machine translation and labelling team for providing the data used for this work. 

\bibliographystyle{IEEEtran}
\bibliography{KWS}

\end{document}